\documentclass[lettersize,journal]{IEEEtran}
\usepackage{amsmath,amsfonts}
\usepackage{algorithmic}
\usepackage{algorithm}
\usepackage{array}
\usepackage[caption=false,font=normalsize,labelfont=sf,textfont=sf]{subfig}
\usepackage{textcomp}
\usepackage{stfloats}
\usepackage{url}
\usepackage{verbatim}
\usepackage{graphicx}
\usepackage{cite}
\usepackage{xcolor}
\usepackage{pifont}
\usepackage{hyperref}

\begin{document}

\title{Generative AI meets 3D: A Survey on Text-to-3D in AIGC Era}

\author{Chenghao Li, Chaoning Zhang,~\IEEEmembership{Senior,~IEEE,} Joseph Cho, Atish Waghwase, Lik-Hang Lee, Francois Rameau, Yang Yang~\IEEEmembership{Senior,~IEEE}, Sung-Ho Bae, Choong Seon Hong,~\IEEEmembership{Fellow,~IEEE}
        % <-this % stops a space

\thanks{Chenghao Li and Atish Waghwase are with the KAIST, South Korea (email: lch17692405449@gmail.com; atishwaghwase@gmail.com).}

\thanks{Chaoning Zhang, Joseph Cho, Sung-Ho Bae and Choong Seon Hong are with the Kyung Hee University, South Korea (email: chaoningzhang1990@gmail.com; joyousaf@khu.ac.kr; shbae@khu.ac.kr; cshong@khu.ac.kr).}

\thanks{Lik-Hang Lee is with the Hong Kong Polytechnic University, Hong Kong SAR (China) (e-mail: lik-hang.lee@polyu.edu.hk).}

\thanks{Francois Rameau is with the State University of New York, Korea, (e-mail: rameau.fr@gmail.com).}

\thanks{Yang Yang is with the University of Electronic Science and Technology, China (e-mail: dlyyang@gmail.com).}}

% The paper headers
\markboth{Journal of \LaTeX\ Class Files,~Vol.~14, No.~8, August~2021}%
{Shell \MakeLowercase{\textit{et al.}}: A Sample Article Using IEEEtran.cls for IEEE Journals}

\maketitle

\begin{abstract}
Generative AI has made significant progress in recent years, with text-guided content generation being the most practical as it facilitates interaction between human instructions and AI-generated content (AIGC). Thanks to advancements in text-to-image and 3D modeling technologies, like neural radiance field (NeRF), text-to-3D has emerged as a nascent yet highly active research field. Our work conducts a comprehensive survey on this topic and follows up on subsequent research progress in the overall field, aiming to help readers interested in this direction quickly catch up with its rapid development. First, we introduce 3D data representations, including both Structured and non-Structured data. Building on this pre-requisite, we introduce various core technologies to achieve satisfactory text-to-3D results. Additionally, we present mainstream baselines and research directions in recent text-to-3D technology, including fidelity, efficiency, consistency, controllability, diversity, and applicability. Furthermore, we summarize the usage of text-to-3D technology in various applications, including avatar generation, texture generation, scene generation and 3D editing. Finally, we discuss the agenda for the future development of text-to-3D.
\end{abstract}

\section{Introduction}

\IEEEPARstart{T}{he} rapid development of generative AI, which drives the creation of AI-generated content (AIGC), has gained significant attention in recent years. The content generation paradigm, guided and constrained by natural language, such as text-to-text (e.g., ChatGPT~\cite{zhang2023one}) and text-to-image~\cite{zhang2023text} (e.g., DALLE-2~\cite{ramesh2022hierarchical}), is the most practical, as it allows for a intuitive interaction between human guidance and generative AI~\cite{zhang2023complete}. The accomplishment of Generative AI in the field of text-to-image~\cite{zhang2023text} is quite remarkable. Given the 3D nature of our environment, we can understand the need to extend this technology to the 3D domain~\cite{ahmed2018survey}, where there is substantial demand for 3D digital content creation across numerous fields and applications~\cite{yang2022fusing} such as including gaming, movies, virtual reality, architecture, and robots, which incorporate tasks such as 3D character generation, 3D texture generation, 3D scene generation, etc~\cite{wang2023survey,kye2021educational}. However, training professional 3D modelers requires extensive artistic and aesthetic training as well as deep technical knowledge~\cite{lv2023generative}. Given the current trends of 3D model development, it is essential to utilize generative AI to produce high-quality and large-scale 3D models~\cite{shi2022deep,lv2023generative}. Furthermore, text-to-3D modeling can significantly aid both novices and professionals in creating 3D content~\cite{jeon2022blockchain}.

\textbf{\textit{Text-to-3D Evolution:}}
% This survey offers a detailed exploration of the evolution of text-to-3D technology, covering its progress from pioneering approaches to the most recent techniques. 
The earliest research on text-to-3D shape generation heavily relied on text-3D pairing data~\cite{chen2019text2shape, achlioptas2019shapeglot}. However, due to their irregular and non-structured properties, 3D shapes pose unique challenges compared to image generation, which makes modeling techniques developed for 2D images hardly applicable.
Aside from this difference, available text-3D dataset~\cite{fu2022shapecrafter} are comparitively smaller than their text-image counterpart~\cite{schuhmann2022laion}. This lack of training data limits variety and volume, hindering model generalization. Many text-3D pairs are also likely synthetic, which may reduce realism. 
The development of Neural Radiance Fields (NeRF)\cite{mildenhall2021nerf}, which generates 3D shapes by rendering views from arbitrary viewpoints, marked a major breakthrough in addressing data scarcity. Advances in multimodal AI\cite{radford2021learning} and diffusion models~\cite{ho2020denoising} further enhanced 3D content creation, improving tools for text-to-3D generation~\cite{jain2022zero,xu2022dream3d,wang2022clip}, though early-stage realism remained limited. More recently, some research has integrated pre-trained image-text models like CLIP~\cite{radford2021learning} with NeRF to enhance 3D shape generation. While these advancements have led to more realistic 3D models, issues with accuracy and realism still persist. To tackle these problems, new methods such as Score Distillation Sampling (SDS)~\cite{poole2022dreamfusion} and Score Jacobian Chains (SJC)~\cite{wang2022score} have been proposed, to improve the quality of generated objects and making them more aligned with text descriptions. These technological advancements have spurred further developments in text-to-3D generation, expanding its applications to include avatars, textures, scenes, and more.

\textbf{\textit{Related Surveys:}} The domain of 3D model generation is extensive, encompassing a wide range of generative representations, methodologies, and scholarly references. Notably, certain studies specifically target the generation of distinct 3D representations, such as 3D point cloud generation~\cite{guo2020deep} and 3D-aware image synthesis~\cite{xia2023survey}. By contrast, multiple comprehensive surveys provide an examination of 3D generation techniques more broadly~\cite{shi2022deep,liu2024comprehensive,li2024advances}. Unlike them, our survey aims to survey the paradigm of generative 3D models from textual descriptions. Moreover, a recent survey~\cite{jiang2024survey}, concurrent to ours but arXived later, has also investigated the specialized area of Text-to-3D. It categorizes text-to-3D generation from a methodological perspective but does not cover its applications. 

By contrast, our survey provides a comprehensive and detailed discussion of seminal text-to-3D methods with five directions for their enhancement and also covers their applications in the major fields, encompassing a broader range of relevant literature.

\textbf{\textit{Scope and structure:}} This survey aims to provide a comprehensive overview of the current state of text-to-3D generation technologies. In Section~\ref{sec:data_representations}. this survey begins by outlining foundational 3D data representation methods, covering both Structured and Non-Structured data representation, such as voxel grids, multi-view images, meshes, point clouds, and neural fields. Section~\ref{sec:basetechs} covers details regarding core technological advances, such as NeRF, diffusion models, etc. Section~\ref{sec:tech} focuses on the seminal text-to-3D methods and summarizes follow-up works that improve them in terms of fidelity, efficiency, consistency, diversity, and controllability. Section~\ref{sec:applications} presents text-to-3D applications, including generation (avatar, scene, and texture) and editing. Section~\ref{sec:discussion} discusses the agenda for future development of text-to-3D.
    
\section{3D Data Representation}\label{sec:data_representations}

3D data has different representations~\cite{minto2018deep,park2019deepsdf,jin20203d,hanocka2019meshcnn,mildenhall2021nerf,lee2024text}, categorized into Structured and non-Structured. Structured data typically has a grid structure. These properties make it relatively straightforward to extend existing 2D deep learning paradigms to 3D data, where convolution operations remain analogous to those in 2D. By contrast, 3D non-structured data lacks a grid structured~\cite{blinn2005pixel}. Therefore, extending classical deep learning techniques to such non-structured representations is a challenging task, which is widely known as geometric deep learning~\cite{cao2020comprehensive} and has attracted significant attention. In the following section, we provide an overview of the major 3D data representations, followed by a brief comparison from three perspectives: representation, computation, and efficiency.

\begin{figure}[h]
    \centering
    \includegraphics[width=\linewidth]{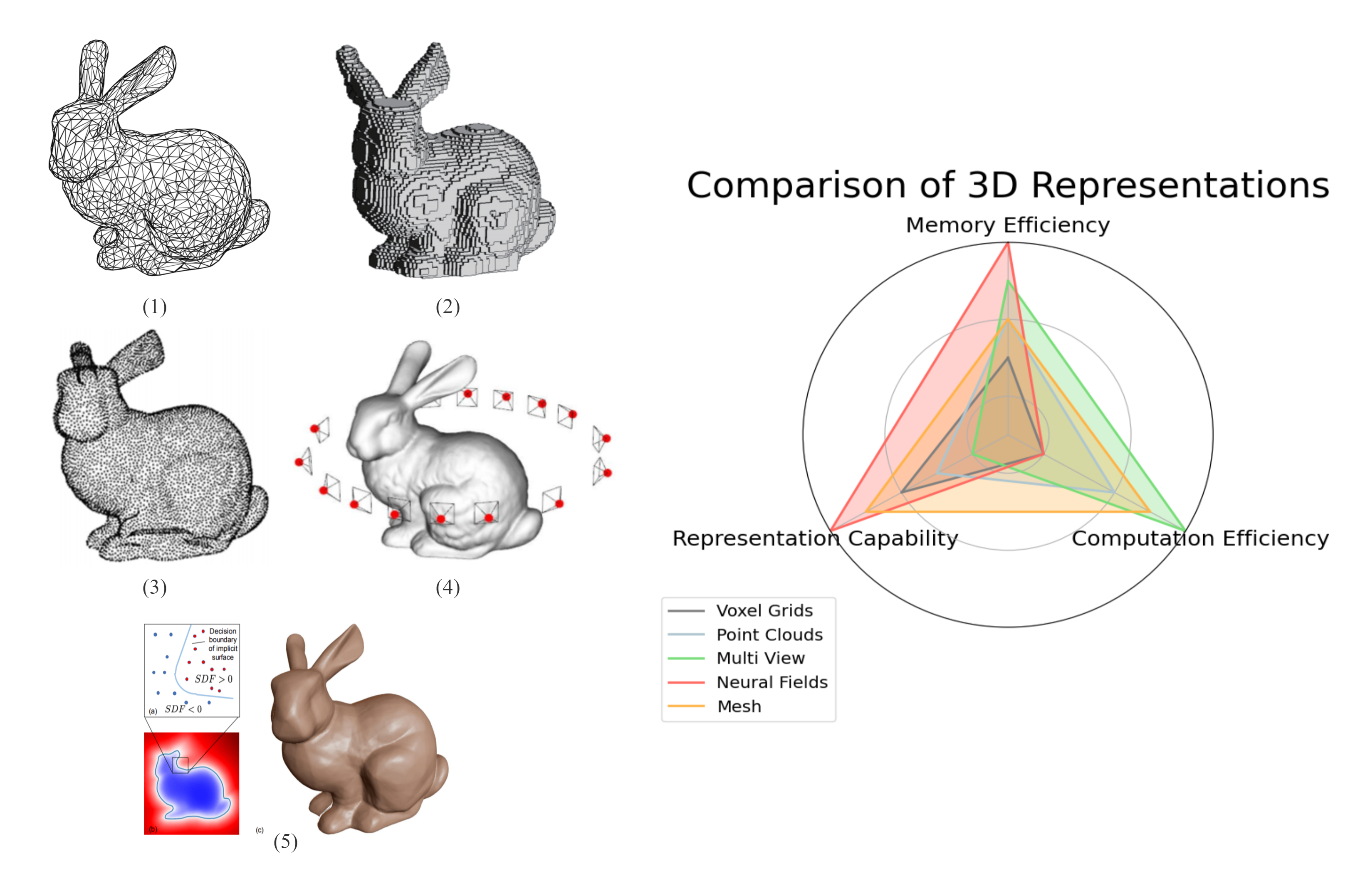}
    \caption{Five 3D representations: (1) mesh, (2) voxel, (3) point cloud, (4) multi-view, and (5) neural field. Representation capability: the capability of a method to capture and express the complexity and details of 3D objects. Computation efficiency: the computation required to generate, render, or process 3D representations. Memory efficiency: the amount of memory required to store 3D representations.}
    \label{fig:voxel}
\end{figure}

\subsection{Structured}

The Structured data preserves the attribute of the grid structure, with global parameterization and a common coordinate system. The major 3D data representations in this category include voxel grids and multi-view images.

\subsubsection{\textbf{Voxel Grids}}

Voxels are individual samples on a regularly spaced 3D grid, akin to pixels in 2D space~\cite{blinn2005pixel}. Each voxel can store single data attributes (e.g., opacity) or multiple attributes (e.g., color and opacity), as well as high-dimensional feature vectors like geometric occupancy~\cite{mescheder2019occupancy}, volumetric density~\cite{minto2018deep}, or signed distance values~\cite{park2019deepsdf}. While a voxel represents a grid point rather than a volume, the space between voxels is unrepresented, leading to information loss that can be reconstructed through interpolation.
Although voxels provide a simple, extensible structure for convolutional neural networks~\cite{wang2019normalnet}, they are inefficient due to large storage requirements from representing both occupied and unoccupied spaces. This limitation makes them less suitable for high-resolution data except in specific applications~\cite{wu2023virtual}. Voxel grids are useful in rendering tasks~\cite{rematas2020neural, hu2023multiscale}, where high-dimensional feature vectors encapsulate the geometry and appearance of scenes, and also in volumetric imaging in medicine and terrain representation in games and simulations.

\subsubsection{\textbf{Multi-view Images}}

A multi-view image dataset consists of multiple posed-images of an object or scene from various perspectives (e.g., front, side, top)~\cite{zhao2018multi}. With advancements in modern digital cameras and computer vision techniques, capturing large quantities of multi-view images has become possible~\cite{cyganek2011introduction}. These multi-view images play an important role in tasks such as multi-view stereo~\cite{furukawa2015multi} and view-consistent image understanding~\cite{dong2022viewfool}. This has created a pressing need to extract 3D structures from these images for applications like 3D reconstruction~\cite{jin20203d}. The main advantage of multi-view datasets is their ability to provide ample data for training 3D models. Although Multi-view Images cannot be strictly defined as 3D data, they serve as a bridge between 2D and 3D visualizations. Moreover, multi-view images are also key training data for NeRF~\cite{mildenhall2021nerf}, effectively meeting the large-scale data requirements of learning-based NeRF methods~\cite{yu2023mvimgnet}.

\subsection{Non-Structured}

In contrast to the Structured 3D data, Non-structured 3D data does not have global parametrization. It mainly consists of three types: 3D meshes, point clouds, and neural fields, which are summarized as follows.

\subsubsection{\textbf{3D Meshes}}

3D meshes are a popular representation of 3D shapes, composed of polygons (faces) defined by vertices that indicate coordinates in 3D space~\cite{wang2022survey} . These vertices are linked by a connectivity list that describes their interconnections. Since meshes only model the surface of a scene, they are more compact and provide connectivity for surface points, making them widely used in traditional computer graphics~\cite{zhou2021three} for tasks like geometry processing, animation, and rendering.
However, meshes are inherently non-structured data, with local geometry represented as subsets of Euclidean space, where properties like shift-invariance and global parameterization are poorly defined, which makes learning on 3D meshes challenging~\cite{ahmed2018survey}. Fortunately, advancements in graph neural networks enable viewing meshes as graphs~\cite{wu2020comprehensive}. For instance, MeshCNN~\cite{hanocka2019meshcnn} introduces convolutional and pooling layers for mesh edges, extracting edge features for shape analysis.
3D meshes are crucial in various fields, including architecture, furniture design, gaming, and medical sciences. 

\subsubsection{\textbf{Point Clouds}}

Point clouds consist of a disordered set of discrete samples representing three-dimensional shapes in space~\cite{qi2017pointnet}. Due to their global non-structured nature, point clouds can also be viewed as globally parametrized small Euclidean subsets, depending on whether the global or local structure is emphasized~\cite{guo2020deep}.
Most applications focus on capturing the object's global characteristics, which reinforces the view of point clouds as non-structured data. Point clouds are directly generated by depth sensors~\cite{liu2019deep}, making them popular in 3D scene understanding tasks. However, their irregularity poses challenges for processing with traditional 2D neural networks.
To address this, various geometric deep learning methods have been developed for analyzing point clouds~\cite{cao2020comprehensive}. For instance, PointNet~\cite{qi2017pointnet} directly processes raw point cloud data and uses sparse keypoints to summarize the input, demonstrating robustness to small perturbations for multiple tasks like shape classification, part segmentation, and scene segmentation.
3D point cloud technology is applicable across multiple fields, including architecture, civil engineering, geological surveys, machine vision, agriculture, space information, and autonomous driving, providing enhanced modeling, analysis, positioning, and tracking accuracy~\cite{qi2017pointnet}.

\subsubsection{\textbf{Neural Fields}}

Neural fields~\cite{peng20043d,park2019deepsdf,shen2021deep,mildenhall2021nerf,muller2022instant,chan2022efficient,gao2022get3d,kerbl20233d,zhang20223dilg,chen2024meshxl} represent scenes or objects in 3D space using neural networks, mapping characteristics to attributes at each point in 3D space. With continuous representation, they can represent 3D scenes or objects at any resolution and complex topology, with early works in this domain focusing on 3D shape representation~\cite{peng20043d}. 
For instance, Signed Distance Functions (SDF)~\cite{park2019deepsdf} use continuous volumetric fields defined by distance and sign at each surface point to represent 3D shapes as neural fields. It has been leveraged in multiple studies~\cite{shen2021deep, gao2022get3d} for shape generation. Neural Radiance Fields (NeRF)~\cite{mildenhall2021nerf} enable high-quality, realistic 3D model generation from any number of input images, requiring no specific processing or labeling. Unlike polygon ray tracing, which needs expensive graphic cards~\cite{gao2022nerf}, neural networks can operate on low-power devices, allowing high-quality rendering on mobile phones and web browsers. 
Hybrid neural fields have also been explored in multiple works to combine explicit and implicit representations. 
For instance, DMTet~\cite{shen2021deep} directly optimizes reconstructed surfaces, synthesizing finer geometric details with fewer artifacts than SDF-based methods, and Triplane~\cite{chan2022efficient} offers fast processing and efficient scalability with increased resolution. Moreover, NeurCF~\cite{chen2024meshxl} provides an explicit coordinate representation with implicit neural embeddings for large-scale sequential mesh modeling.
The above advantages come at the cost of high computational resource demands~\cite{gao2022nerf,hu2023multiscale}, difficulties in handling complex scenes and lighting~\cite{gao2022nerf}, which poses challenges for their direct applications in 3D assets, like VR and gaming~\cite{mildenhall2021nerf}. Nevertheless, the emerging neural fields represent a promising 3D representation.

\noindent\textit{\textbf{A Brief Comparison:}} We present a brief comparison of the above five 3D representation types in terms of their representation capability, computation efficiency, and memory efficiency (see Fig.~\ref{fig:voxel}). \textit{(a)} Neural Fields and Mesh demonstrate high representational capability, accurately capturing complex geometric shapes and details~\cite{gao2022nerf}. Voxel Grids are also competitive in representation, making them suitable for describing intricate structures. By contrast, Point Clouds show relatively weak performance when dealing with details and areas of lower density~\cite{cao2020comprehensive}. Multi View is has the lowest representational capability, as it struggles to precisely reconstruct complex geometric forms~\cite{furukawa2015multi}. \textit{(b)} In terms of computational efficiency, however, multi View performs the best as it only needs to process images from multiple perspectives. Mesh has advantages due to its relatively simple structure~\cite{wang2022survey}, allowing for quick rendering and processing. Point Clouds have a moderate level of computational efficiency since they require certain point cloud processing algorithms but still demand more computation.  Voxel Grids and Neural Fields perform poorly because they consume a significant amount of computational resources when handling high-resolution data and consume a large amount of computational resources when handling neural network computations~\cite{blinn2005pixel}. \textit{(c)} For memory efficiency, Neural Fields performs the best because they compress 3D information through neural networks. Multi View follows closely, with a storage format of image sequences, resulting in low memory usage~\cite{yu2023mvimgnet}. Point Clouds and Meshes are average since storing sparse points and polygonal meshes requires more space. By contrast, Voxel Grids require the most memory at high resolutions due to their uniform grid format~\cite{blinn2005pixel}. Overall, neural fields are relatively more promising with their competitive representation capability and memory efficiency, and the advantage of being computation-heavy is expected to be overcome with the advancement of GPUs and other computation tools.

\section{Foundation Technologies}~\label{sec:basetechs}

In this section, we introduce those foundation technologies that contribute to the modern text-to-3D methods. Specifically, we briefly summarize Neural Radiance Field, Diffusion models, and the other two representative technologies, with a timeline of these technologies presented in Figure~\ref{fig:TimeLine}.

\subsection{\textbf{Neural Radiance Field}}

Neural Radiance Field (NeRF)~\cite{mildenhall2021nerf, gao2022nerf} is a neural network-based implicit representation of 3D scenes, which can render synthetic images from arbitrary viewpoints and a given position. Specifically, given a 3D point $\mathbf{x} \in \mathbb{R}^3$ and an observation direction unit vector $\mathbf{d}\in \mathbb{R}^2$~\cite{mildenhall2021nerf}, NeRF encodes the scene as a continuous volumetric radiance field $f$, yielding a differential density $\sigma$ and an RGB color $\mathbf{c}$: $f(\mathbf{x}, \mathbf{d}) = (\sigma, \mathbf{c})$. Rendering of images from desired perspectives can be achieved by integrating color along a suitable ray, $\mathbf{r}$, for each pixel in accordance with the volume rendering equation~\cite{mildenhall2021nerf} shown as:

\begin{equation}
    \hat{\mathcal{C}}(r) = \int_{t_n}^{t_f} T(t)\sigma(t)\mathbf{c}(t) dt,   
\end{equation}

\begin{equation}
    T(t) = exp\left(-\int_{t_n}^{t}\sigma(s)ds\right).
\end{equation}
The transmission coefficient $T(t)$ is defined as the probability that light is not absorbed from the near-field boundary $t_n$ to $t$.
In order to train NeRF network and optimize the predicted color $\hat{\mathcal{C}}$ to fit with the ray $\mathcal{R}$ corresponding to the pixel in the training images, gradient descent is used to optimize the network and match the target pixel color by loss~\cite{gao2022nerf}:

\begin{equation}
    \mathcal{L}=\sum_{\mathbf{r} \in \mathcal{R}}\|\mathcal{C}(\mathbf{r})-\mathcal{\hat{C}}(\mathbf{r})\|_2^2
\end{equation}\\

\subsection{\textbf{Diffusion Model}}

The use of diffusion model~\cite{ho2020denoising} has seen a dramatic increase in the past few years. Known as denoising diffusion probabilistic models (DDPMs) or score-based generative models, these models generate new data that is similar to the data used to train them~\cite{dhariwal2021diffusion}. Drawing inspiration from non-equilibrium thermodynamics, DDPMs are defined as a parameterized Markov chain of diffusion steps that adds random noise to the training data and learns to reverse the diffusion process to produce the desired data samples from the pure noise~\cite{song2020score}. In the forward process, DDPM destroys the training data by gradually adding Gaussian noise. It starts from a data sample $\mathbf{x}_0$ and iteratively generates noisier samples $\mathbf{x}_T$ with $q(\mathbf{x}_t\mid\mathbf{x}_{t-1})$, using a Gaussian diffusion kernel:

\begin{align}
q(x_{1:T} | x_0) := \prod_{t=1}^T q( x_t | x_{t-1} ), \label{eq:forwardprocess_1}
\end{align}
\begin{align}
q(x_t|x_{t-1}) := \mathcal{N}(x_t;\sqrt{1-\beta_t} x_{t-1},\beta_t I) \label{eq:forwardprocess_2},
\end{align}
where $T$ and $\beta_t$ are the steps and hyper-parameters, respectively. We can obtain noised image at arbitrary step $t$ with Gaussian noise transition kernel as $\mathcal{N}$ in Eq.~\ref{eq:forwardprocess_2}, by setting $\alpha_t := 1 - \beta_t$ and $\bar{\alpha}_t := \prod_{s=0}^{t} \alpha_s$:

\begin{align}
q(x_t|x_{0}) := \mathcal{N}(x_t;\sqrt{\bar{\alpha}_t}x_{0},(1 - \bar{\alpha}_t) I) \label{eq:forwardprocess_3}
\end{align}
The reverse denoising process of DDPM involves learning to undo the forward diffusion by performing iterative denoising, thereby generating data from random noise~\cite{ho2020denoising}. This process is formally defined as a stochastic process, with an optimization objective is to generate $p_\theta(x_0)$ which follows the true data distribution $q(x_0)$ by starting from $p_\theta(T)$:

\begin{align}
     E_{t \sim \mathcal{U}( 1,T ), \mathbf x_0 \sim q(\mathbf x_0), \epsilon \sim \mathcal{N}(\mathbf{0},\mathbf{I})}{ \lambda(t)  \left\| \epsilon - \epsilon_\theta(\mathbf{x}_t, t) \right\|^2} \label{eq:loss}
\end{align}
The integration of diffusion models with 3D data has shown promising results~\cite{bautista2022gaudi} in tasks such as 3D shape generation and scene reconstruction~\cite{shue20233d}, demonstrating their disruptive potential in the field of 3D content creation.

\subsection{\textbf{Other Representative Technologies}}
Neural field and diffusion models form the foundation of modern text-to-3D methods; however, other technologies are also essential, and we highlight two representative ones. 

\textbf{Unified Text-Vision Representation.} Given the text-guided nature, a unified representation of text and vision is necessary for realizing text-to-3D. CLIP~\cite{radford2021learning} (Contrastive Language-Image Pre-training) is a seminal work for providing such a unified representation. 
CLIP employs a symmetric InfoNCE loss~\cite{oord2018representation} to jointly train image and text encoders, enabling the prediction of correct pairings from a batch of (image, text) training samples. By projecting images and text into a shared vector space, CLIP facilitates the mutual mapping of semantic information between images and text.
When combined with diffusion models~\cite{ho2020denoising}, pre-trained text-visual unified models can be utilized for diffusion-based text-to-image generation, exemplified by models such as GLIDE~\cite{nichol2021glide} and Imagen~\cite{saharia2022photorealistic}, which then becomes a valuable prior for text-to-3D generation~\cite{zhang2023text}.
To ensure that the generated 3D models are consistent with text descriptions from various perspectives, text-to-3D techniques typically render models from multiple angles and employ unified text and visual models to compute matching scores for each viewpoint~\cite{poole2022dreamfusion}. By leveraging unified text and visual models, text-to-3D techniques are capable of generating high-quality 3D models, enhancing expressiveness and semantic consistency~\cite{wang2022score}.

\textbf{Score Distillation Sampling.} Text-to-3D generation through diffusion models and neural fields typically relies on SDS (Score Distillation Sampling)~\cite{poole2022dreamfusion} optimization or a similar variant termed SJC (Score Jacobian Chaining)~\cite{wang2022score}. It comprises two essential components: a neural field representation, akin to the 3D model like NeRF, and a pretrained text-to-image diffusion model~\cite{poole2022dreamfusion}. The 3D model produces images $x$ at specified camera positions, which can be expressed as a parametric function $x = g(\theta)$, with  $g$ denoting the chosen volumetric renderer and $\theta$ representing a coordinate-based neural network. The diffusion model~\cite{ho2020denoising} $\phi$ features a learned denoising function $\epsilon \phi(x_t; y, t)$, which forecasts sampled noise $\epsilon$ derived from the noisy image $x_t$, noise level $t$, and text embedding $y$. This denoising function supplies the gradient direction for updating $\theta$, aiming to steer all rendered images towards high-probability density regions conditioned on the text embedding under the diffusion prior~\cite{poole2022dreamfusion}:

\begin{align}
    \nabla_{\theta} L_{\text{SDS}}(\phi, g(\theta)) = \mathbb{E}_{t,\epsilon} \left( w(t) (\epsilon \phi(x_t; y, t) - \epsilon) \frac{\partial x}{\partial \theta} \right).
\end{align}
In this context, $w(t)$ represents a weighting function, while both the scene model $g$ and the diffusion model $\phi$ serve as modular elements within the framework, offering flexibility in selection. By gradually updating the parameters of the generator in the direction of the gradient, SDS can generate images that better match the given text prompts~\cite{poole2022dreamfusion}.

\begin{figure*}
    \centering
    \includegraphics[width=\textwidth]{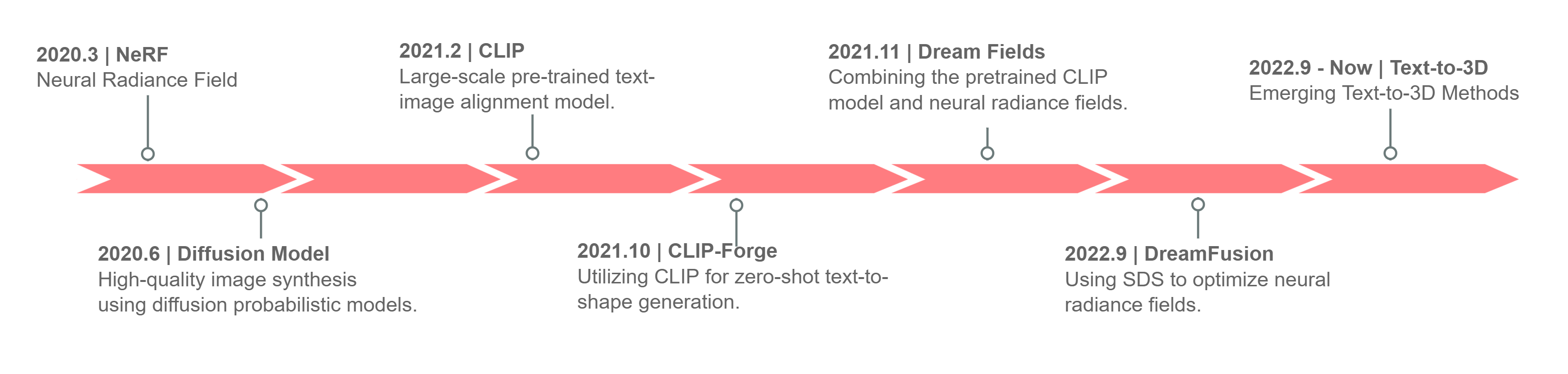}
    \caption{Timeline of foundation technologies that contribute to modern text-to-3D methods.}
    \label{fig:TimeLine}
\end{figure*}

\section{Text-to-3D Methods and Directions}\label{sec:tech}

Built on top of the above foundational technologies, some works have proposed seminal text-to-3D methods. Moreover, numerous follow-up works have emerged to address the limitations of those seminal methods~\cite{jain2022zero,mohammad2022clip,poole2022dreamfusion,lin2022magic3d,wang2022score,xu2022dream3d,han2023zero3d}. This section first introduces those seminal methods and then summarizes new directions for their enhancement, with an overview presented in Figure~\ref{fig:TimeLine}. 

\subsection{Seminal Methods}

With the success of text-to-image generation modeling~\cite{zhang2023text}, text-to-3D generation has also gained attention from the deep learning community~\cite{jain2022zero,mohammad2022clip,poole2022dreamfusion,lin2022magic3d,wang2022score,xu2022dream3d,han2023zero3d}. However, the scarcity of 3D data makes expanding datasets challenging. Early approaches such as Dream Fields~\cite{jain2022zero} and CLIP-Mesh~\cite{mohammad2022clip} first rely on pre-trained image-text models~\cite{radford2021learning} to optimize underlying 3D representations (RMS and meshes) in order to alleviate the training data problem, achieving high text-image alignment scores for all 2D renderings. Although these methods avoid the costly requirement of 3D training data and primarily rely on large-scale pre-trained image-text models, they often produce unrealistic 2D renderings. To address these limitations, DreamFusion~\cite{poole2022dreamfusion} and SJC~\cite{wang2022score} utilize powerful pre-trained text-to-image diffusion models as strong image priors and employ neural fields as the 3D representation, showcasing impressive capabilities in text-to-3D synthesis. The development of text-to-3D generation are featured by three seminal methods: \textit{CLIP-Forge}~\cite{sanghi2022clip}, \textit{Dream Fields}~\cite{jain2022zero}, and \textit{DreamFusion}~\cite{poole2022dreamfusion}, which are summarized in the following. 

To begin with, CLIP-Forge\cite{sanghi2022clip} is the first to introduce a pre-trained text-to-image model into 3D generation tasks, successfully achieving text-to-shape conversion without the need for inference time optimization through an efficient generation process. This not only improves generation efficiency, but also demonstrates the model's flexibility in generating diverse shapes. Following this, Dream Fields~\cite{jain2022zero} further expands the field by using CLIP to synthesize and manipulate 3D object representations. Specifically, it combines neural rendering and multimodal representations to generate diverse 3D objects from natural language descriptions, significantly enhancing the visual quality and multi-view consistency of the generated results. Additionally, DreamFusion~\cite{poole2022dreamfusion} adopts a similar approach to Dream Fields but replaces CLIP with a 2D diffusion model and employs optimization strategies to enhance generation outcomes. Its optimization-based training approach allows for significant improvements in the fidelity and stability of the generated 3D shapes, showcasing the powerful capabilities of generative models~\cite{poole2022dreamfusion}.

\begin{figure*}
    \centering
    \includegraphics[width=\linewidth]{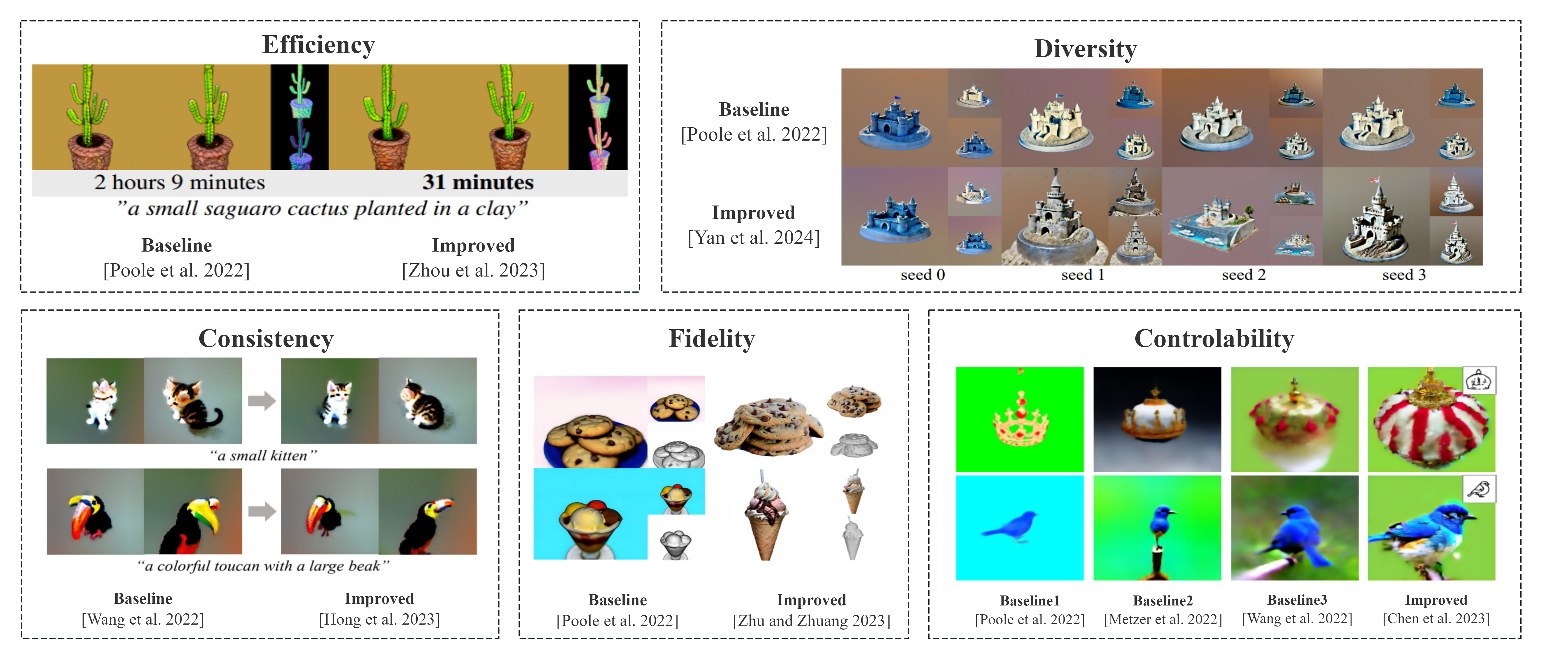}
    \caption{Five enhancement cases represent improvements in fidelity, diversity, consistency, efficiency, and controllability. \textbf{Efficiency} shows time reduction from 2 hours 9 minutes~\cite{poole2022dreamfusion} to 31 minutes~\cite{zhou2023dreampropeller}. \textbf{Diversity} demonstrates more varied sandcastle models, improving from~\cite{poole2022dreamfusion} to~\cite{yan2024flow}. \textbf{Consistency} shows better visual coherence in generating a kitten and toucan, evolving from~\cite{wang2022score} to~\cite{hong2023debiasing}. \textbf{Fidelity} highlights improved image quality, from~\cite{poole2022dreamfusion} to~\cite{zhu2023hifa}. \textbf{Controllability} displays finer control in generated crowns and birds, advancing from~\cite{poole2022dreamfusion,metzer2022latent,wang2022score} to~\cite{chen2023control3d}.
    }
    \label{fig:Enhancement}
\end{figure*}

It is worth mentioning other notable works that have played a pioneering role in the development of text-to-3D generation. For instance, \textit{CLIP-Sculptor}\cite{sanghi2022clip} employs image-shape pairs as supervision and adopts a multi-resolution, voxel-based conditioned generation scheme along with discrete latent representations to generate diverse 3D shapes. Moving further, \textit{Latent-NeRF}\cite{metzer2022latent} incorporates both textual guidance and shape guidance for image and 3D model generation, as well as a latent-dispersal model for direct application of dispersed rendering on 3D meshes. Lastly, a novel framework for editing 3D object styles based on textual description, has been introduced in\textit{Text2Mesh}~\cite{michel2022text2mesh} for handling low-quality meshes without the need for pre-trained generative models or specialized 3D mesh datasets.

\begin{table*}[h]
    \centering
    \caption{List of directions to enhance text-to-3d methods.}
    \scalebox{1}{
    \begin{tabular}{lccccc}
    \hline
    Method & Project & Representation & Optimization & Directions \\
    \hline
    3D-CLFusion~\cite{li20233d} & - & NeRF & Diffusion+Contrast & Efficiency \\
    3DTopia~\cite{hong20243dtopia} & \href{https://github.com/3DTopia/3DTopia}{Link} & Triplane & SDS & Efficiency \\
    3DFuse~\cite{seo2023let} & \href{https://ku-cvlab.github.io/3DFuse/}{Link} & NeRF & SDS/SJC & Consistency\&Controllability \\
    ATT3D~\cite{lorraine2023att3d} & \href{https://research.nvidia.com/labs/toronto-ai/ATT3D/}{Link} & NeRF & SDS & Efficiency \\
    CLIP-NeRF~\cite{wang2022clip} & \href{https://cassiepython.github.io/clipnerf/}{Link} & NeRF & CLIP & Controllability \\
    Consist3D~\cite{ouyang2023chasing} & - & SJC & SDS+sem+warp+rec & Consistency \\
    Consistent3D~\cite{wu2024consistent3d} & \href{https://github.com/sail-sg/Consistent3D}{Link} & NeRF/DMTet/3DGS & CDS & Consistency \\
    Control3D~\cite{chen2023control3d} & - & NeRF & C-SDS & Controllability \\
    CorrespondentDream~\cite{kim2024enhancing} & - & NeRF & SDS & Fidelity \\
    CSD~\cite{yu2023text} & \href{https://xinyu-andy.github.io/Classifier-Score-Distillation}{Link} & NeRF/DMTet & CSD & Fidelity \\
    DATID-3D~\cite{kim2023datid} & \href{https://gwang-kim.github.io/datid_3d/}{Link} & Triplane & ADA+den & Diversity \\
    Diffusion-SDF~\cite{li2023diffusion} & \href{https://github.com/ttlmh/Diffusion-SDF}{Link} & SDF & Diffusion-SDF & Diversity \\
    DITTO-NeRF~\cite{seo2023ditto} & \href{https://janeyeon.github.io/ditto-nerf}{Link} & NeRF & inpainting-SDS & Efficiency\&Consistency\&Diversity \\
    D-SDS~\cite{hong2023debiasing} & \href{https://susunghong.github.io/Debiased-Score-Distillation-Sampling/}{Link} & NeRF & Debiased-SDS & Consistency \\
    Dream3D~\cite{xu2022dream3d} & \href{https://bluestyle97.github.io/dream3d/}{Link} & DVGO & CLIP+prior & Controllability \\
    DreamBooth3D~\cite{raj2023dreambooth3d} & \href{https://dreambooth3d.github.io/}{Link} & NeRF & SDS+MVR & Controllability \\
    DreamCraft3D~\cite{sun2023dreamcraft3d} & \href{https://github.com/deepseek-ai/DreamCraft3D}{Link} & NeuS+DMTet & BSD & Consistency \\
    DreamGaussian~\cite{tang2023dreamgaussian} & \href{https://dreamgaussian.github.io/}{Link} & 3DGS & SDS & Efficiency \\
    DreamPropeller~\cite{zhou2023dreampropeller} & \href{https://github.com/alexzhou907/DreamPropeller}{Link} & NeRF/DMTet & SDS/VSD & Efficiency \\
    DreamTime~\cite{huang2023dreamtime} & - & NeRF & TP-VSD/TP-SDS & Fidelity\&Diversity \\
    Dreamer XL~\cite{miao2024dreamer} & \href{https://github.com/xingy038/Dreamer-XL}{Link} & 3DGS & TSM & Consistency \\
    EfficientDreamer~\cite{zhao2023efficientdreamer} & \href{https://efficientdreamer.github.io/}{Link} & NeuS+DMTet & SDS/VDS & Consistency \\
    ExactDreamer~\cite{zhang2024exactdreamer} & \href{https://github.com/zymvszym/ExactDreamer}{Link} & 3DGS & ESM & Fidelity\&Consistency \\
    Fantasia3D~\cite{chen2023fantasia3d} & \href{https://fantasia3d.github.io/}{Link} & DMTet & SDS & Fidelity \\
    FSD~\cite{yan2024flow} & - & NeRF & FSD & Diversity \\
    GaussianDiffusion~\cite{li2023gaussiandiffusion} & - & 3DGS & SDS & Fidelity\&Consistency \\
    GaussianDreamer~\cite{yi2023gaussiandreamer} & \href{https://taoranyi.com/gaussiandreamer/}{Link} & 3DGS & SDS & Efficiency \\
    GSGEN~\cite{chen2023text} & \href{https://gsgen3d.github.io/}{Link} & 3DGS & SDS & Efficiency\&Consistency \\
    Grounded-Dreamer~\cite{li2024grounded} & - & NeRF & SDS & Fidelity \\
    HD-Fusion~\cite{wu2024hd} & - & SDF+DMTet & SDS/VSD & Fidelity \\
    HiFA~\cite{zhu2023hifa} & \href{https://hifa-team.github.io/HiFA-site/}{Link} & NeRF & SDS & Fidelity\&Consistency \\
    InNeRF360~\cite{wang2024innerf360} & \href{https://ivrl.github.io/InNeRF360/}{Link} & NeRF & DSDS & Consistency\&Controllability \\
    Instant3D~\cite{li2023instant3d} & \href{https://jiahao.ai/instant3d/}{Link} & Triplane & MSE+LPIPS & Efficiency\&Diversity \\
    Interactive3D~\cite{dong2024interactive3d} & \href{https://interactive-3d.github.io/}{Link} & 3DGS+InstantNGP & interactive-SDS & Controllability \\
    IT3D~\cite{chen2024it3d} & \href{https://github.com/buaacyw/IT3D-text-to-3D}{Link} & NeRF+Mesh & SDS & Fidelity\&Consistency \\
    LI3D~\cite{lin2023towards} & - & NeRF & SDS & Controllability \\
    LucidDreamer~\cite{liang2023luciddreamer} & \href{https://github.com/EnVision-Research/LucidDreamer}{Link} & 3DGS & ISM & Fidelity \\
    Magic3D~\cite{lin2022magic3d} & \href{https://research.nvidia.com/labs/dir/magic3d}{Link} & NeRF+DMTet & SDS & Fidelity\&Efficiency \\
    MATLABER~\cite{xu2023matlaber} & \href{https://sheldontsui.github.io/projects/Matlaber}{Link} & DMTet & SDS & Fidelity \\
    MTN~\cite{yi2023progressive} & - & Multi-Scale Triplane & SDS & Fidelity \\
    MVControl~\cite{li2023mvcontrol} & \href{https://github.com/WU-CVGL/MVControl}{Link} & NeuS/DMTet & SDS & Controllability \\
    MVDream~\cite{shi2023mvdream} & \href{https://mv-dream.github.io/}{Link} & NeRF & SDS & Consistency \\
    Perp-Neg~\cite{armandpour2023re} & \href{https://perp-neg.github.io/}{Link} & NeRF & SDS & Consistency \\
    PI3D~\cite{liu2023pi3d} & - & Triplane & SDS & Efficiency\&Consistency \\
    Points-to-3D~\cite{yu2023points} & - & NeRF & SDS & Consistency\&Controllability \\
    ProlificDreamer~\cite{wang2024prolificdreamer} & \href{https://ml.cs.tsinghua.edu.cn/prolificdreamer/}{Link} & NeRF & VSD & Fidelity\&Diversity \\
    RichDreamer~\cite{qiu2023richdreamer} & \href{https://aigc3d.github.io/richdreamer/}{Link} & NeRF/DMTet & SDS & Consistency \\
    Sherpa3D~\cite{liu2023sherpa3d} & \href{https://liuff19.github.io/Sherpa3D/}{Link} & DMTet & SDS & Consistency \\
    SweetDreamer~\cite{li2023sweetdreamer} & \href{https://sweetdreamer3d.github.io/}{Link} & NeRF/DMTet & SDS & Consistency \\
    TAPS3D~\cite{wei2023taps3d} & \href{https://github.com/plusmultiply/TAPS3D}{Link} & DMTet & CLIP+IMG & Fidelity\&Diversity \\
    TextMesh~\cite{tsalicoglou2023textmesh} & \href{https://fabi92.github.io/textmesh/}{Link} & SDF+Mesh & SDS & Fidelity \\
    X-Dreamer~\cite{ma2023x} & \href{https://xmu-xiaoma666.github.io/Projects/X-Dreamer}{Link} & DMTet & SDS+AMA & Fidelity \\
    X-Oscar~\cite{ma2024x} & \href{https://xmu-xiaoma666.github.io/Projects/X-Oscar/}{Link} & SMPL-X & ASDS & Fidelity \\
     % ~\cite{} & \href{}{Link} &  &  & Fidelity\&Efficiency\&Consistency\&Controllability\&Diversity\&Applicability \\
    \hline
    \end{tabular}
    }
    \label{tab:enhanced_text-to-3d}
\end{table*}

\subsection{New Directions}

As summarized above, the text-to-3D generation paradigm that combines NeRF and text-to-image-prior is a promising research direction. However, some challenging issues remain, such as low fidelity, long inference time, consistency issues, poor controllability, and low diversity. Such remaining issues are alleviated by numerous follow-up works (See Table~\ref{tab:enhanced_text-to-3d}).

\subsubsection{\textbf{Fidelity}}

The 3D generation model should closely resemble the actual object in terms of shape, texture, lighting, and other aspects, which require high fidelity. Constrained by the weakly supervised and low-resolution CLIP, the upscaled results are often not satisfactory, which is reflected in the general fidelity of the generated models~\cite{xu2022dream3d}. Fidelity and speed often involve trade-offs, where improving inference speed can come at the cost of fidelity~\cite{poole2022dreamfusion}. Films require high-precision models, while games often require more quantity than film-level precision~\cite{wang2023survey}.
Multiple methods have been developed to optimize fidelity. 
\textit{Dream3D}\cite{xu2022dream3d} introduces an explicit 3D shape prior into the CLIP-guided optimization process, improving shape accuracy relative to input text. Building on this, \textit{HD-Fusion}\cite{wu2024hd} enhances quality by integrating multiple noise estimation processes with a pretrained 2D diffusion prior, showcasing varied strategies for common challenges;
In parallel, \textit{MTN}\cite{yi2023progressive} employs a multi-scale triplane network and progressive learning for detail recovery, complementing the approaches of \textit{HD-Fusion}. The work on \textit{CSD} (Classifier Score Distillation)\cite{yu2023text} emphasizes the role of classifier scores in optimizing score-based diffusion synthesis (SDS), reinforcing the focus on fidelity shared by \textit{Dream3D} and \textit{HD-Fusion};
Additionally, \textit{X-Dreamer}\cite{ma2023x} tackles the domain gap in text-to-3D creation with Camera-Guided Low-Rank Adaptation and Attention-Mask Alignment Loss, paralleling the coherence issues addressed by \textit{HD-Fusion}. In contrast, \textit{LucidDreamer}\cite{liang2023luciddreamer} improves quality by mitigating excessive smoothing through Interval Score Matching (ISM), while \textit{HiFA}~\cite{zhu2023hifa} introduces a single-stage optimization method to overcome artifacts;
\textit{Grounded-Dreamer}\cite{li2024grounded} enhances fidelity with attention refocusing, while \textit{CorrespondentDream}\cite{kim2024enhancing} uses cross-view correspondences. \textit{ExactDreamer}\cite{zhang2024exactdreamer} employs Exact Score Matching (ESM) to correct reconstruction errors and improve detail generation.

\subsubsection{\textbf{Efficiency}}
A fatal issue of generating 3D content by leveraging pre-training models based on diffusion models as a powerful prior and learning objective is that the inference process is too slow. Even at a resolution of just 64$\times$64, 
For instance, both \textit{Magic3D}\cite{lin2022magic3d} and \textit{3D-CLFusion}\cite{li20233d} tackle time issues effectively. \textit{Magic3D} employs a two-phase optimization framework, starting with a low-resolution diffusion prior and followed by a sparse 3D hash grid structure for acceleration. In contrast, \textit{3D-CLFusion} utilizes a pre-trained latent NeRF, achieving fast 3D content creation in under a minute;
Similarly, \textit{ATT3D}\cite{lorraine2023att3d} and \textit{PI3D}\cite{liu2023pi3d} focus on time reduction through innovative frameworks. \textit{ATT3D} trains multiple prompts on a unified model, while \textit{PI3D} employs lightweight iterative refinement and score-based diffusion synthesis (SDS) to generate high-quality outputs within just 3 minutes;
Furthermore, \textit{DreamGaussian}\cite{tang2023dreamgaussian} and \textit{GaussianDreamer}\cite{yi2023gaussiandreamer} enhance mesh generation speed. \textit{DreamGaussian} produces textured meshes from single-view images in 2 minutes using a 3D Gaussian splatting model, whereas \textit{GaussianDreamer} accelerates model generation by integrating 2D and 3D diffusion models;
Lastly, \textit{DreamPropeller}\cite{zhou2023dreampropeller} achieves up to a 4.7x speedup in any text-to-3D pipeline based on score distillation, maintaining high quality. Both \textit{3D-CLFusion} and its view-invariant diffusion approach leverage contrastive learning\cite{he2020momentum} for rapid generation.

\subsubsection{\textbf{Consistency}}
In 3D scenes generated by models such as DreamFusion~\cite{poole2022dreamfusion}, distortion and ghost are often observed. To remove them, the generation model should maintain a consistent shape and feature regardless of the viewing angle. 
Additionally, unstable 3D scenes are often observed when text prompts or random seeds are changed, which is mainly caused by the lack of perception of 3D information from 2D prior diffusion models. The transmission model has no knowledge of which direction the object is observed from, which leads to the serious distortion of 3D scenes by generating the front-view geometry features from all viewpoints, including the sides and the back~\cite{poole2022dreamfusion}. Various techniques have been introduced to resolve such inconsistency.
Hong \textit{et al.}~\cite{hong2023debiasing} has introduced two debiasing methods: score debiasing, which gradually increases the truncation value of the 2D diffusion model's estimation, and prompt debiasing, which uses a language model to align user prompts with view prompts.
Meanwhile, \textit{3DFuse}\cite{seo2023let} optimizes the training of the 2D diffusion model to effectively process sparse 3D structures while ensuring semantic consistency across viewpoints. Additionally, \textit{Perp-Neg}\cite{armandpour2023re} leverages geometrical properties of the score space to address the Janus problem in text-to-image diffusion models, enhancing generation flexibility. Moreover, \textit{EfficientDreamer}\cite{zhao2023efficientdreamer} utilizes orthogonal-view diffusion priors to improve the fidelity of generated 3D models, while \textit{MVDream}\cite{shi2023mvdream} presents a diffusion model that generates consistent multiview images from text prompts, bridging 2D and 3D data. \textit{DreamCraft3D}~\cite{sun2023dreamcraft3d} achieves high-fidelity 3D object generation using 2D reference images and view-dependent diffusion models;
Furthermore, \textit{Sherpa3D}\cite{liu2023sherpa3d} employs coarse 3D prior-guided strategies to refine prompts, addressing the Janus problem, while \textit{Consistent3D}\cite{wu2024consistent3d} introduces deterministic trajectory sampling priors to mitigate geometry collapse and poor texture quality in score-based diffusion synthesis (SDS). Lastly, \textit{Dreamer XL}~\cite{miao2024dreamer} implements Trajectory Score Matching (TSM) to resolve pseudo ground truth inconsistencies caused by accumulated errors during the Denoising Diffusion Implicit Model (DDIM) inversion process.

\subsubsection{\textbf{Controllability}}
While text-to-3D models can generate impressive results, they often remain unconstrained, leading to challenges like guiding collapse. High controllability is desired to let the user freely adjust the attributes of generated 3D objects according to specific requirements, such as shape, color, and style. Poor controllability, however, has long been a challenge in text-to-3D generation tasks. To address this, ControlNet~\cite{zhang2023adding} has added extra input conditions, such as canny edges, hough lines, and depth maps, to provide more control over the generation process. Such a unique combination of text and shape guidance represents a significant step towards enhancing precision in 3D content generation. Building on this, multiple methods have further refined control mechanisms.
For instance, \textit{CLIP-NeRF}\cite{wang2022clip} enables intuitive interaction with NeRF using the CLIP model's language-image embedding space. Similarly, \textit{TAPS3D}\cite{wei2023taps3d} simplifies generation with pseudo captions, removing the need for additional optimization.
To enhance input flexibility, \textit{Control3D}\cite{chen2023control3d} integrates hand-drawn sketches to guide NeRF learning, allowing nuanced control over 3D content. Likewise, \textit{MVControl}\cite{li2023mvcontrol} builds on pre-trained multi-view 2D diffusion models to enable controllable multi-view image generation.
Additionally, \textit{Interactive3D}\cite{dong2024interactive3d} employs a two-stage approach for direct user interaction. Lastly, \textit{3DFuse}\cite{seo2023let} enhances robustness and consistency by constructing a rough 3D structure from text prompts and using a projected depth map.

\subsubsection{\textbf{Diversity}}

A limitation of text-to-3D methods is the lack of sample diversity in the adapted generative models, primarily due to the deterministic nature of the text encoder. In the context of 3D generation, achieving diversity is more challenging, as it requires extensive datasets of training images along with their associated camera distribution information~\cite{kim2023datid}.
Another key issue that affects diversity in 3D generation lies in the distillation objectives of Score Distillation Sampling (SDS) methods. These methods are designed to maximize the likelihood of generating images from 3D representations. This often results in overly consistent outputs, reducing the variety of generated models.
\begin{comment}

\end{comment}
Meveral works have been introduced to address these issues. \textit{Diffusion-SDF}~\cite{li2023diffusion} combines an SDF autoencoder with a voxelized diffusion model to generate more diverse 3D shapes. \textit{DATID-3D}~\cite{kim2023datid} utilizes text-to-image diffusion models to generate diverse images from text prompts without requiring additional images or camera information for the target domain. \textit{Instant3D}~\cite{li2023instant3d} improves the rapid generation of high-diversity 3D assets through fine-tuning SDXL~\cite{podell2023sdxl} and a transformer-based reconstruction model. \textit{Flow Score Distillation (FSD)}~\cite{yan2024flow} significantly enhances diversity while maintaining quality by improving the noise sampling strategy. Finally, \textit{DITTO-NeRF}\cite{seo2023ditto} introduces a progressive 3D object reconstruction scheme, including scale, orientation, and masks, leading to improvements in diversity.\\

\noindent It is worth noting that some works simultaneously target the aforementioned multiple aspects. For instance, \textit{Magic3D}~\cite{lin2022magic3d} introduces a two-stage optimization framework to enhance NeRF for improved speed and resolution. \textit{Points-to-3D}~\cite{yu2023points} achieves view-consistent and shape-controllable text-to-3D generation by introducing sparse 3D points. Moreover, \textit{GSGEN}~\cite{chen2023text} combines Gaussian Splatting with a progressive optimization strategy to address challenges like inaccurate geometry and slow speed in text-to-3D methods. Meanwhile, \textit{GaussianDiffusion}\cite{li2023gaussiandiffusion} utilizes variational Gaussian splatting for precise image saturation control and multi-view consistency. \textit{InNeRF360}\cite{wang2024innerf360} employs depth-space warping to maintain consistency in multiview segmentations, refining the NeRF model with perceptual and geometric priors. 

\section{Text-to-3D applications} \label{sec:applications}

With the methods getting increasingly mature for generating 3D content, numerous text-to-3D applications have emerged in various fields, including text-guided 3D Avatar Generation, Scene Generation, Texture Generation, and 3D editing.

\subsection{\textbf{Text Guided 3D Avatar Generation}}

Skinned Multi-Person Linear (SMPL)~\cite{loper2023smpl} is a widely-used skeleton-driven parametric human model that deforms 3D meshes using shape and pose components. It provides two gender-specific template meshes and uses PCA bases and rotation matrices to describe body shape and pose. Due to its simplicity and open-source nature, It has been adopted in numerous works for 3D avatar generation, including 3D objects and human features like hair, facial expressions, and clothing~\cite{lv2023generative,yang2022fusing}.

Early works~\cite{cao2023dreamavatar,zhao2023zero,zhang2023dreamface,jiang2023avatarcraft,tevet2022motionclip,hong2022avatarclip,wang2023rodin} explore combining text-based 2D priors with neural fields for 3D avatar generation. Recent works focus on generating full-body avatars~\cite{cao2023dreamavatar,jiang2023avatarcraft,tevet2022motionclip,hong2022avatarclip,tu2023motioneditor}.
\textit{DreamAvatar}\cite{cao2023dreamavatar} utilizes a combination of trainable NeRF and a text-to-image diffusion model to create 3D avatars with fine-tuned poses, employing the SMPL\cite{bogo2016keep} model for precise shape control. In a similar approach, \textit{AvatarCraft}~\cite{jiang2023avatarcraft} addresses character identity and style generation by integrating a diffusion model, enhanced by multi-boundary box strategies for detailed texture and geometry creation.
Building on avatar generation techniques, methods like \textit{MotionCLIP}\cite{tevet2022motionclip} and \textit{AvatarCLIP}\cite{hong2022avatarclip} focus on motion synthesis by aligning 3D motion auto-encoders with CLIP's latent space, enabling text-driven animation. While \textit{MotionCLIP} targets motion interpolation and editing, \textit{AvatarCLIP} excels in generating avatars and animations in zero-shot scenarios, broadening the scope of text-driven 3D content creation.
For motion editing, \textit{MotionEditor}~\cite{tu2023motioneditor} offers a distinct approach by employing a dual-branch architecture, maintaining the original scene’s integrity while refining motion dynamics. Collectively, these advancements illustrate a trend towards integrating multi-modal guidance and neural representations, pushing the boundaries of both avatar realism and animation flexibility in text-to-3D generation. Some works focus on generating head avatars~\cite{zhao2023zero,zhang2023dreamface,wang2023rodin}, such as \textit{T2P}\cite{zhao2023zero}, which leverages CLIP and neural rendering to search for both continuous and discrete facial parameters within a unified framework, enabling zero-shot text-driven automatic creation of game characters. Another work, \textit{DreamFace}\cite{zhang2023dreamface} introduces a progressive scheme for personalized 3D facial generation guided by text, allowing users to customize 3D facial assets with desired shapes, textures, and animation capabilities. 
% It employs a coarse-to-fine scheme, SDS, and a dual-path mechanism for neutral appearance generation, enhancing animation capability and rendering quality. 
Lastly, \textit{Rodin}~\cite{wang2023rodin} preserves the integrity of 3D diffusion while providing much-needed computational efficiency, demonstrating the ability to generate 3D digital avatars from text and allowing text-guided editing.\\

\begin{table*}[h]
    \centering
    \caption{List of text-guided 3D avatar generation methods.}
    \scalebox{1}{
    \begin{tabular}{lcccccc}
    \hline
    Method & Project & Generation Part & Motivation & Animateable  & Editable \\
    \hline
    AvatarBooth~\cite{zeng2023avatarbooth} & \href{https://zeng-yifei.github.io/avatarbooth_page/}{Link} & Full-Body & Fidelity\&Consistency & \checkmark & \checkmark \\
    AvatarCLIP~\cite{hong2022avatarclip} & \href{https://github.com/hongfz16/AvatarCLIP}{Link} & Full-Body & Diversity & \checkmark & - \\
    AvatarCraft~\cite{jiang2023avatarcraft} & \href{https://avatar-craft.github.io/}{Link} & Full-Body & Controllability & \checkmark & \checkmark \\
    AvatarFusion~\cite{huang2023avatarfusion} & \href{https://hansenhuang0823.github.io/AvatarFusion}{Link} & Full-Body\&Clothes & Diversity & \checkmark & \checkmark \\
    AvatarVerse~\cite{zhang2024avatarverse} & \href{https://avatarverse3d.github.io/}{Link} & Full-Body & Fidelity & \checkmark & \checkmark \\
    Chupa~\cite{kim2023chupa} & \href{https://snuvclab.github.io/chupa}{Link} & Full-Body & Fidelity\&Diversity & - & \checkmark \\
    DreamAvatar~\cite{cao2023dreamavatar} & - & Full-Body & Fidelity & - & \checkmark \\
    DreamFace~\cite{zhang2023dreamface} & \href{https://sites.google.com/view/dreamface}{Link} & Head & Fidelity\&Diversity & \checkmark & \checkmark \\
    DreamHuman~\cite{kolotouros2024dreamhuman} & \href{https://dream-human.github.io/}{Link} & Full-Body & Fidelity\&Diversity & \checkmark & - \\
    DreamWaltz~\cite{huang2024dreamwaltz} & \href{https://idea-research.github.io/DreamWaltz/}{Link} & Full-Body & Fidelity\&Diversity & - & \checkmark \\
    DressCode~\cite{he2024dresscode} & \href{https://ihe-kaii.github.io/DressCode/}{Link} & Clothes & Fidelity\&Diversity & - & \checkmark \\
    GenerateCT~\cite{hamamci2023generatect} & \href{https://github.com/ibrahimethemhamamci/GenerateCT}{Link} & Chest & Medicine Applicability & - & - \\
    HeadArtist~\cite{liu2023headartist} & \href{https://kumapowerliu.github.io/HeadArtist}{Link} & Head & Fidelity\&Controllability & - & \checkmark\\
    HeadSculpt~\cite{han2024headsculpt} & \href{https://brandonhan.uk/HeadSculpt}{Link} & Head & Fidelity\&Controllability & - & \checkmark \\
    HeadStudio~\cite{zhou2024headstudio} & - & Head & Fidelity\&Efficiency & \checkmark & - \\
    HumanNorm~\cite{huang2023humannorm} & \href{https://humannorm.github.io/}{Link} & Full-Body & Fidelity & \checkmark & \checkmark \\
    HumanGaussian~\cite{liu2023humangaussian} & \href{https://alvinliu0.github.io/projects/HumanGaussian}{Link} & Full-Body & Fidelity\&Efficiency & - & - \\
    Human4DiT~\cite{shao2024human4dit} & \href{https://human4dit.github.io/}{Link} & Full-Body & Quality\&Consistency & \checkmark & - \\
    LAGA~\cite{gong2024laga} & \href{https://gongjia0208.github.io/LAGA/}{Link} & Full-Body\&Clothes & Fidelity\&Diversity & - & \checkmark \\
    Make-It-Vivid~\cite{tang2024make} & \href{https://make-it-vivid.github.io/}{Link} & Full-Body & Fidelity\&Diversity & \checkmark & \checkmark \\
    MotionCLIP~\cite{tevet2022motionclip} & \href{https://guytevet.github.io/motionclip-page/}{Link} & Motion & Controllability & \checkmark & \checkmark \\
    MotionEditor~\cite{tu2023motioneditor} & \href{https://francis-rings.github.io/MotionEditor/}{Link} & Motion & Controllability & \checkmark & \checkmark\\
    PBRGAN~\cite{wang2024text} & - & Head & Fidelity\&Efficiency\&Diversity & - & - \\
    Portrait3D~\cite{wu2024portrait3d} & - & Portrait & Fidelity\&Consistency & - & - \\
    Rodin~\cite{wang2023rodin} & \href{https://3d-avatar-diffusion.microsoft.com/}{Link} & Head & Fidelity\&Efficiency\&Consistency & - & \checkmark \\
    SEEAvatar~\cite{xu2023seeavatar} & \href{https://yoxu515.github.io/SEEAvatar/}{Link} & Full-Body & Fidelity & - & - \\
    T2P~\cite{zhao2023zero} & - & Head & Diversity & - & \checkmark \\
    TADA~\cite{liao2023tada} & \href{https://tada.is.tue.mpg.de/}{Link} & Full-Body & Fidelity\&Diversity\&Consistency & \checkmark & - \\
    TECA~\cite{zhang2023text} & \href{https://yfeng95.github.io/teca/}{Link} & Head & Fidelity & - & \checkmark \\
    TeCH~\cite{huang2023tech} & \href{https://huangyangyi.github.io/TeCH}{Link} & Full-Body & Fidelity\&Controllability & \checkmark & \checkmark \\
    X-Oscar~\cite{ma2024x} & \href{https://xmu-xiaoma666.github.io/Projects/X-Oscar/}{Link} & Full-Body & Fidelity & \checkmark & - \\
    % ~\cite{} & \href{}{Link} &  &  & \checkmark & \\
    \hline
    \end{tabular}
    }
    \label{tab:comparison}
\end{table*}

Realistic and diverse 3D human model generation has been a recent focus~\cite{kim2023chupa,kolotouros2024dreamhuman,huang2023avatarfusion,zhang2024avatarverse,huang2023humannorm,huang2024dreamwaltz,xu2023seeavatar,ma2024x,gong2024laga,tang2024make,wu2024portrait3d}. 
\textit{Chupa}\cite{kim2023chupa} focuses on enhancing identity diversity through 2D normal maps to guide 3D avatar reconstruction. In contrast, \textit{DreamHuman}\cite{kolotouros2024dreamhuman} integrates neural radiance fields and statistical models to expand both diversity and fidelity in text-to-image synthesis. Moving towards quality improvements, \textit{AvatarFusion}\cite{huang2023avatarfusion} and \textit{AvatarVerse}\cite{zhang2024avatarverse} employ dual volume rendering and progressive high-resolution synthesis, respectively, to address view consistency and avatar detail.
Multiple works have specifically targeted texture and geometry fidelity. \textit{HumanNorm}\cite{huang2023humannorm} uses a normal diffusion model alongside SDS loss for precision in geometry, while \textit{DreamWaltz}\cite{huang2024dreamwaltz} leverages 3D-consistent occlusion-aware SDS to enhance both fidelity and diversity. Similarly, \textit{SEEAvatar}\cite{xu2023seeavatar} employs self-evolving constraints to refine appearance quality. Tackling oversaturation, \textit{X-Oscar}\cite{ma2024x} introduces Adaptive Variational Parameter (AVP) and Avatar-aware Score Distillation Sampling (ASDS) for improved generation clarity.
For avatar customization, methods like \textit{LAGA}\cite{gong2024laga} utilize Gaussian point layers to generate detailed clothing, while \textit{Make-It-Vivid}\cite{tang2024make} enhances textures with adversarial learning informed by vision-question-answering agents. Finally, \textit{Portrait3D}~\cite{wu2024portrait3d} addresses Janus effects and oversaturation using a joint geometry-appearance prior and a pyramid tri-grid representation, rounding out the strategies for ensuring high-fidelity avatar synthesis.

3D avatar generation relies on high controllability and animation capability of 3D human model generation~\cite{zeng2023avatarbooth,han2024headsculpt,pavlakos2019expressive,liao2023tada,zhang2023text,kerbl20233d,zhou2024headstudio,huang2024dreamwaltz,gong2024laga}. To this end, \textit{AvatarBooth}\cite{zeng2023avatarbooth} enhances control and model accuracy through pose-consistency constraints and a multi-resolution rendering strategy. \textit{HeadSculpt}\cite{han2024headsculpt} achieves fidelity and editability via landmark-based control and text-embedding learning, introducing an identity-aware editing score distillation strategy. Leveraging SMPL-X~\cite{pavlakos2019expressive}, \textit{TADA}\cite{liao2023tada} enhances geometric-texture consistency for animatable character generation. \textit{TECA}\cite{zhang2023text} seamlessly transfers composite features between avatars, supporting powerful editing effects. Utilizing 3D Gaussian splatting~\cite{kerbl20233d}, \textit{HeadStudio}\cite{zhou2024headstudio} generates realistic and animated digital avatars. Enabling the creation of complex shapes and appearances, as well as new poses for animation, \textit{DreamWaltz}\cite{huang2024dreamwaltz} facilitates the development of 3D avatars. Finally, \textit{LAGA}~\cite{gong2024laga} permits convenient garment-level editing by decoupling clothing from the avatar.

\subsection{\textbf{Text Guided 3D Scene Generation}}

3D scenes play a crucial role in fields such as video production, gaming, and the metaverse, ~\cite{lv2023generative}. However, generating 3D scenes still faces significant challenges, requiring large-scale scene reconstruction, multi-view images, and ensuring the realism and consistency of the scenes. 3D scene modeling is also a time-consuming task, usually requiring professional 3D designers to complete. Recently, some efforts have attempted to address such challenges.\\

\noindent\textbf{Outward-Facing.} The most common type of generated visuals is outward-facing. "Outward-facing" typically refers to the direction or angle facing away from the interior. In the context of 3D scene generation, outward-facing often refers to the perspective observed or captured from inside the scene towards the exterior, providing the viewpoint seen by an external observer. Outward-facing viewpoints are crucial for simulating observation and navigation in the real world. Some works primarily generate these perspectives~\cite{hollein2023text2room,zhang2023text2nerf,song2023roomdreamer,fang2023ctrl,li2023generative}. 
\textit{Text2Room}\cite{hollein2023text2room} breaks new ground by generating room-scale 3D meshes with textures from text, focusing on entire environments rather than single objects\cite{poole2022dreamfusion,lin2022magic3d} or trajectory scaling as in SceneScape~\cite{fridman2023scenescape}. Expanding on this, \textit{Text2NeRF}~\cite{zhang2023text2nerf} combines NeRF with diffusion models to achieve zero-shot 3D scene generation from text, using a progressive inpainting strategy and multi-view constraints for consistent, diverse scenes. It introduces depth-aware NeRF optimization, tackling view alignment issues with a two-stage depth correction.
Building on these advancements, \textit{RoomDreamer}\cite{song2023roomdreamer} aligns scene synthesis with structural prompts through Geometry Guided Diffusion for consistent styling and Mesh Optimization to refine geometry and texture. Similarly, \textit{Ctrl-Room}\cite{fang2023ctrl} enables interactive scene editing by first learning layout distributions and then generating detailed panoramas via a fine-tuned ControlNet~\cite{zhang2023adding}, enhancing both visual coherence and editability.
In contrast, \textit{FastScene}~\cite{li2023generative} prioritizes speed and scene quality with Coarse View Synthesis and Progressive Novel View Inpainting, using Multi-View Projection and 3D Gaussian Splatting to streamline reconstruction. This approach not only accelerates generation but also improves scene realism, positioning it as a alternative to existing methods. There are also some other types of 3D scene generation perspectives, such as Object-Centered~\cite{po2023compositional,singer2023text}, Perpetual View~\cite{fridman2023scenescape,li2024art3d,lee2024vividdream}, and Object-Compositional~\cite{lin2023componerf,cohen2023set,rahamim2024lay}.

\begin{table*}[h]
    \centering
    \caption{List of text-guided 3D scene generation methods.}
    \scalebox{1}{
    \begin{tabular}{lcccc}
    \hline
    Method & Project & View Type & Motivation & Editable\\
    \hline
    ART3D ~\cite{li2024art3d} & - & Perpetual View & Consistency & - \\
    CLIP3Dstyler~\cite{gao2023clip3dstyler} & - & Outward-Facing & Controllability\&Diversity & - \\
    CompoNeRF~\cite{lin2023componerf} &\href{https://github.com/hbai98/Componerf}{Link} & Object-Compositional &  Consistency\&Controllability & \checkmark \\
    Ctrl-Room~\cite{fang2023ctrl} & - & Outward-Facing & Consistency\&Controllability & \checkmark \\
    FastScene~\cite{ma2024fastscene} & - & Outward-Facing & Fidelity\&Efficiency\&Consistency & - \\
    GALA3D~\cite{zhou2024gala3d} & \href{https://gala3d.github.io/}{Link} & Object-Centered & Fidelity\&Controllability & \checkmark \\
    Lay-A-Scene~\cite{rahamim2024lay} & \href{https://lay-a-scene.github.io/}{Link} & Object-Compositional & Consistency\&Controllability & \checkmark \\
    MAV3D~\cite{singer2023text} & \href{https://make-a-video3d.github.io/}{Link} & Object-Centered & Fidelity\&Consistency & - \\
    RoomDreamer~\cite{song2023roomdreamer} & - & Outward-Facing & Fidelity\&Consistency & \checkmark \\
    SceneScape~\cite{fridman2023scenescape} & \href{https://scenescape.github.io/}{Link} & Perpetual View & Consistency\&Diversity & - \\
    Set-the-Scene~\cite{cohen2023set} & - & Object-Compositional & Controllability & \checkmark \\
    TC4D~\cite{bahmani2024tc4d} & \href{https://sherwinbahmani.github.io/tc4d/}{Link} & Object-Centered & Fidelity\&Controllability & - \\
    Text2NeRF~\cite{zhang2023text2nerf} & \href{https://github.com/eckertzhang/Text2NeRF}{Link} & Outward-Facing & Fidelity\&Consistency\&Diversity & - \\
    Text2Room~\cite{hollein2023text2room} & \href{https://lukashoel.github.io/text-to-room}{Link} & Outward-Facing & Consistency\&Diversity & \checkmark \\
    VividDream~\cite{lee2024vividdream} & \href{https://vivid-dream-4d.github.io/}{Link} & Perpetual View & Consistency\&Controllability & - \\
    4D-fy~\cite{bahmani20244d} & \href{https://sherwinbahmani.github.io/4dfy/}{Link} & Object-Centered & Fidelity & - \\
    Po~\textit{et al.}~\cite{po2023compositional} & - & Object-Centered & Fidelity\&Efficiency\&Controllability & \checkmark \\
    % ~\cite{} & \href{}{Link} & Object-Centered & Fidelity\&Efficiency\&Controllability & \checkmark \\
    \hline
    \end{tabular}
    }
    \label{tab:texture}
\end{table*}

Object-Centered perspective focuses on adjusting the camera's position and orientation relative to specific objects in a scene, enhancing the viewer's focus on those objects. \textit{MAV3D} (Make-A-Video3D)~\cite{singer2023text} advances 3D dynamic scene generation from text using a 4D dynamic NeRF framework, leveraging Text-to-Video (T2V)~\cite{singer2022make} diffusion-based optimization to enhance scene appearance and motion consistency. By eliminating the need for explicit 3D or 4D data, it outperforms prior baselines in dynamic video creation. Extending the focus to text-to-4D generation, \textit{4D-fy}~\cite{bahmani20244d} employs hybrid score distillation sampling, integrating multiple diffusion models to achieve high-fidelity results. Similarly, \textit{TC4D}~\cite{bahmani2024tc4d} introduces trajectory conditioning to refine control over entity motion within compositional 4D scenes. In contrast, for static 3D synthesis, Po's work~\cite{po2023compositional} utilizes a local condition diffusion approach, enabling precise control over scene components with text hints and bounding boxes. Enhancing compositional optimization further, \textit{GALA3D}~\cite{zhou2024gala3d} focuses on object-scene consistency, producing realistic 3D scenes with coherent geometry, texture, and object interactions.

Perpetual View allows continuous exploration of a scene without limitations, enhancing the understanding of its dynamic characteristics.
\textit{SceneScape}\cite{fridman2023scenescape} introduces a text-driven method for generating long videos of various scenes based solely on text inputs. It combines a pre-trained text-to-image model\cite{rombach2022high} with geometry priors from a monocular depth prediction model~\cite{ranftl2021vision,ranftl2020towards}, achieving 3D consistency through online training and enabling diverse scene generation, such as walking through a spaceship or an ice city.
\textit{ART3D}~\cite{li2024art3d} merges diffusion models with 3D Gaussian splatting to create high-quality artistic scenes. It uses an image semantic transfer algorithm to bridge artistic and realistic images while generating a point cloud map and enhancing 3D scene consistency with a depth consistency module.
\textit{VividDream}~\cite{lee2024vividdream} generates explorable 4D scenes with ambient dynamics from a single image or text prompt, expanding the input into a static 3D point cloud and creating a dynamic video ensemble for perpetual view exploration.

Object-Compositional perspective involves observing objects as complex structures made up of multiple components, aiding in understanding their arrangements and interactions in 3D scenes.
\textit{CompoNeRF}~\cite{lin2023componerf} enables flexible editing and recombination of trained local NeRFs into new scenes using 3D layout manipulation or textual hints, producing faithful and editable text-to-3D results while facilitating multi-object composition.
\textit{Set-the-Scene}~\cite{cohen2023set} introduces an agent-based global-local training framework for synthesizing 3D scenes, allowing the creation of harmonious scenes with style and lighting while learning complete representations of each object. It supports various editing options, such as adjusting placement or deleting objects.
\textit{Lay-A-Scene}~\cite{rahamim2024lay} leverages pre-trained text-to-image models to arrange unseen 3D objects, generating coherent and feasible 3D scenes.

\subsection{\textbf{Text Guided 3D Texture Generation}}

Although text-to-image generation has made rapid progress, creating 3D objects remains a significant challenge because it requires consideration of the specific shape of the surface being rendered~\cite{richardson2023texture}. Fully automated 3D content generation is still constrained by the laborious human efforts required to design textures. Therefore, automating the texture design process through text has become an intriguing yet challenging research problem~\cite{cao2023texfusion}. Synthesized textures not only need to align closely with the textual prompts but also must exhibit high-quality and consistent characteristics across the target mesh. Recently, there have been a number of works on text-to-texture~\cite{richardson2023texture,chen2022tango,ma2023x,chen2023text2tex,gao2023clip3dstyler,cao2023texfusion,wang2024text}.

\begin{table}[h]
    \centering
    \caption{List of text-guided 3D texture generation methods.}
    \scalebox{0.9}{
    \begin{tabular}{lccc}
    \hline
    Method & Project & Entity & Motivation\\
    \hline
    CLIP3Dstyler~\cite{gao2023clip3dstyler} & - & Scene & Controllability\&Diversity \\
    TANGO~\cite{chen2022tango} & \href{https://cyw-3d.github.io/tango/}{Link} & General & Fidelity\\
    TexFusion~\cite{cao2023texfusion} & \href{https://research.nvidia.com/labs/toronto-ai/texfusion/}{Link} & General & Controllability\\
    TEXTure~\cite{richardson2023texture} & \href{https://texturepaper.github.io/TEXTurePaper/}{Link} & General & Controllability\\
    Text2Tex~\cite{chen2023text2tex} & \href{https://daveredrum.github.io/Text2Tex/}{Link} & General & Consistency\\
    X-Mesh~\cite{ma2023x} & \href{https://xmu-xiaoma666.github.io/Projects/X-Mesh/}{Link} & General & Efficiency\\
    PBRGAN~\cite{wang2024text} & - & Face & Fidelity\&Efficiency\&Diversity \\
    % ~\cite{} & \href{}{Link} & Avatar\&Object &  \\
    \hline
    \end{tabular}
    }
    \label{tab:texture}
\end{table}

\textit{TANGO}\cite{chen2022tango} employs the CLIP model to decompose 3D style into reflectance properties, geometric variations, and lighting conditions. This achieves realistic 3D style transfer on arbitrary topology surface meshes and is even suitable for low-quality meshes. Similarly, another study, \textit{TexFusion}\cite{cao2023texfusion}, introduces a 3D consistency generation technique that leverages large-scale text guidance to efficiently generate high-quality and globally consistent textures. In a similar vein, \textit{TEXTure}\cite{richardson2023texture} utilizes a pre-trained deep-to-image topology model to generate seamless 3D textures from different viewpoints and supports texture editing and transfer. Additionally, \textit{Text2Tex}\cite{chen2023text2tex} integrates pre-trained models to address inconsistencies and stretching artifacts in text-driven texture generation, which progressively produce high-resolution textures. Furthermore, Ma \textit{et al.} proposed \textit{X-Mesh}~\cite{ma2023x}, a text-driven 3D stylization framework featuring a Text-Guided Dynamic Attention Module (TDAM), which enables more accurate attribute prediction and faster convergence. Certain techniques focus on texture generation for specific entities. For instance, \textit{PBRGAN}\cite{wang2024text} employs a progressive latent space refinement technique to automatically generate high-quality 3D facial textures, enhancing GANs' capability in diverse texture generation and supporting multi-view consistency. \textit{CLIP3Dstyler}\cite{gao2023clip3dstyler} combines point cloud and text feature matching to achieve 3D scene stylization, enhancing style distinctiveness.

\subsection{\textbf{Text Guided 3D Editing}}

\begin{table*}[h]
    \centering
    \caption{List of text-guided 3D editing methods.}
    \scalebox{1}{
    \begin{tabular}{lccccc}
    \hline
    Method & Project & Add. Guidance & Area & Type & Motivation\\
    \hline
    CLIP3Dstyler~\cite{gao2023clip3dstyler} & - & - & Global & Style & Controllability\&Diversity \\
    ClipFace~\cite{aneja2023clipface} & \href{https://shivangi-aneja.github.io/projects/clipface/}{Link} & - & Global & Texture & Controllability\&Diversity \\
    CompoNeRF~\cite{lin2023componerf} &\href{https://github.com/hbai98/Componerf}{Link} & Layout & User-Defined ROI & Shape\&Texture &  Consistency\&Controllability \\
    Control4D~\cite{shao2023control4d} & \href{https://control4darxiv.github.io/}{Link} & - & Global & Shape\&Texture & Consistency\&Efficiency\&Controllability \\
    DreamEditor~\cite{zhuang2023dreameditor} & \href{https://www.sysu-hcp.net/projects/cv/111.html}{Link} & - & Model-Defined ROI & Shape\&Texture & Consistency\&Controllability \\
    ED-NeRF~\cite{park2023ed} & \href{https://jhq1234.github.io/ed-nerf.github.io/}{Link} & - & Model-Defined ROI & Shape\&Texture & Efficiency\&Controllability \\
    FusionDeformer~\cite{xu2024fusiondeformer} & - & - & Global & Shape &  Fidelity\&Controllability \\
    FocalDreamer~\cite{li2024focaldreamer} & \href{https://focaldreamer.github.io/}{Link} & Region & User-Defined ROI & Shape\&Texture & Consistency\&Controllability \\
    GaussianEditor~\cite{fang2023gaussianeditor} & \href{https://gaussianeditor.github.io/}{Link} & - & Model-Defined ROI & Shape\&Texture & Efficiency\&Controllability \\
    Instruct 3D-to-3D~\cite{kamata2023instruct} & \href{https://sony.github.io/Instruct3Dto3D-doc/}{Link} & Image & Global & Shape\&Texture & Fidelity\&Consistency\&Controllability \\
    InstructP2P~\cite{xu2023instructp2p} & - & - & Global & Shape\&Color & Controllability \\
    Instruct-NeRF2NeRF~\cite{haque2023instruct} & \href{https://instruct-nerf2nerf.github.io/}{Link} & - & Model-Defined ROI & Shape\&Texture & Consistency\&Controllability \\
    InNeRF360~\cite{wang2023inpaintnerf360} & \href{https://ivrl.github.io/InNeRF360}{Link} & - & Model-Defined ROI & Shape\&Texture & Consistency\&Controllability \\
    Progressive3D~\cite{cheng2023progressive3d} & \href{https://cxh0519.github.io/projects/Progressive3D/}{Link} & Region & User-Defined ROI & Shape\&Texture & Controllability\&Applicability \\
    SketchDream~\cite{liu2024sketchdream} & - & Sketch & User-Defined ROI & Shape\&Texture & Consistency\&Controllability \\
    SKED~\cite{mikaeili2023sked} & \href{https://sked-paper.github.io/}{Link} & Sketch & User-Defined ROI & Shape\&Texture & Controllability \\
    TIP-Editor~\cite{zhuang2024tip} & \href{https://zjy526223908.github.io/TIP-Editor/}{Link} & Image & User-Defined ROI & Shape\&Texture & Controllability\&Applicability \\
    TextDeformer~\cite{gao2023textdeformer} & - & - & Global & Shape & Consistency\&Controllability \\
    Vox-E~\cite{sella2023vox} & \href{http://vox-e.github.io/}{Link} & - & Model-Defined ROI & Shape\&Texture & Controllability\&Applicability \\
    Chen~\textit{et al.}~\cite{chen2024text} & \href{https://text-mesh-refinement.github.io/}{Link} & - & Global & Shape\&Texture & Consistency\&Controllability \\
    % ~\cite{} & \href{}{Link} &  &  & Shape\&Texture &  \\
    \hline
    \end{tabular}
    }
    \label{tab:texture}
\end{table*}

\noindent\textbf{Global Editing.} Some works achieve simple and effective global editing through text~\cite{aneja2023clipface,kamata2023instruct,gao2023textdeformer,shao2023control4d,xu2023instructp2p}. 
For instance, \textit{ClipFace}~\cite{aneja2023clipface} is a self-supervised method for text-guided 3D facial texture editing. It uses adversarial training and differentiable rendering with the CLIP model. After training, it predicts both facial textures and expression parameters.
Similarly, \textit{Instruct 3D-to-3D}\cite{kamata2023instruct} transforms 3D models using a pre-trained image-to-image diffusion model. It enhances 3D consistency with dynamic scaling and explicit conditioning on the input 3D scene, surpassing baseline methods\cite{poole2022dreamfusion,wang2022clip}.
\textit{TextDeformer}\cite{gao2023textdeformer} is guided entirely by text prompts to generate deformations on input triangle meshes. It relies on pre-trained image encoders like CLIP\cite{radford2021learning} and DINO\cite{caron2021emerging} to generate large, low-frequency shape changes as well as small, high-frequency details. To overcome issues, TextDeformer proposes using the Jacobian matrix to represent mesh deformation and encourages computing deep features on 2D encoded rendering to ensure consistency.
\textit{Control4D}~\cite{shao2023control4d} edits dynamic 4D portraits with text instructions. It introduces GaussianPlanes and a 4D generator to enhance editing efficiency and consistency.
\textit{InstructP2P}~\cite{xu2023instructp2p} edits 3D point clouds using text guidance. It combines a point cloud diffusion model and a language model to edit color and geometry, showing strong generalization to new shapes.
Lastly, Chen~\textit{et al.}~\cite{chen2024text} adds geometric details to coarse 3D meshes using a text prompt. They apply single-view preview and multi-view normal generation for fast and precise editing.\\

\noindent\textbf{User-Defined Local Editing.} Some works perform user-specified local editing through text and additional input guidance.
% ~\cite{lin2023componerf,mikaeili2023sked,cheng2023progressive3d,li2024focaldreamer,liu2024sketchdream}.
CompoNeRF~\cite{lin2023componerf} introduces an innovative framework that combines editable 3D scene layouts to tackle guidance collapse in text-to-3D generation, enabling flexible editing and recombination of local NeRFs for multi-object composition. In parallel, SKED~\cite{mikaeili2023sked} enhances user interaction with a sketch-based technique that allows intuitive editing of 3D shapes directly from user sketches.
Progressive3D~\cite{cheng2023progressive3d} follows this trend by decomposing the generation process into locally progressive editing steps, focusing changes on user-defined regions to effectively manage complex 3D tasks.
FocalDreamer~\cite{li2024focaldreamer} further refines this approach by facilitating fine-grained editing through the merging of base shapes and customizable parts, employing geometric union and dual-path rendering for high-fidelity outputs while maintaining consistency through innovative loss functions.
In addition, SketchDream~\cite{liu2024sketchdream} supports NeRF generation from hand-drawn sketches and enables localized editing, thereby integrating user creativity into the text-to-3D pipeline.\\

\noindent\textbf{Model-Defined Local Editing.} Some works involve models automatically selecting areas of interest for local editing based on the text.

\textit{Instruct-NeRF2NeRF}\cite{haque2023instruct} introduces a method for text-guided editing of NeRF scenes using an iterative image-based diffusion model (InstructPix2Pix)\cite{brooks2022instructpix2pix} to achieve realistic modifications in large-scale scenarios. In contrast, \textit{Vox-E}~\cite{sella2023vox} employs latent diffusion models to refine 3D objects through volumetric regularization and cross-attention grids.
Building on user-controlled editing, \textit{DreamEditor}\cite{zhuang2023dreameditor} enables localized neural field editing via text prompts, ensuring consistency with a pretrained text-to-image diffusion model. \textit{GaussianEditor}\cite{fang2023gaussianeditor} enhances local editing precision with 3D Gaussians, aligning edits with text instructions for faster training.
For efficiency, \textit{ED-NeRF}\cite{park2023ed} embeds scenes into the latent space of a diffusion model, introducing DDS distillation to speed up editing while maintaining quality. \textit{InNeRF360}\cite{wang2023inpaintnerf360} tackles object removal in 360° Neural Radiance Fields by automating content filling using pre-trained NeRF. 

\section{Discussion}~\label{sec:discussion}
\begin{figure}[!htbp]
    \centering
    \includegraphics[width=\linewidth]{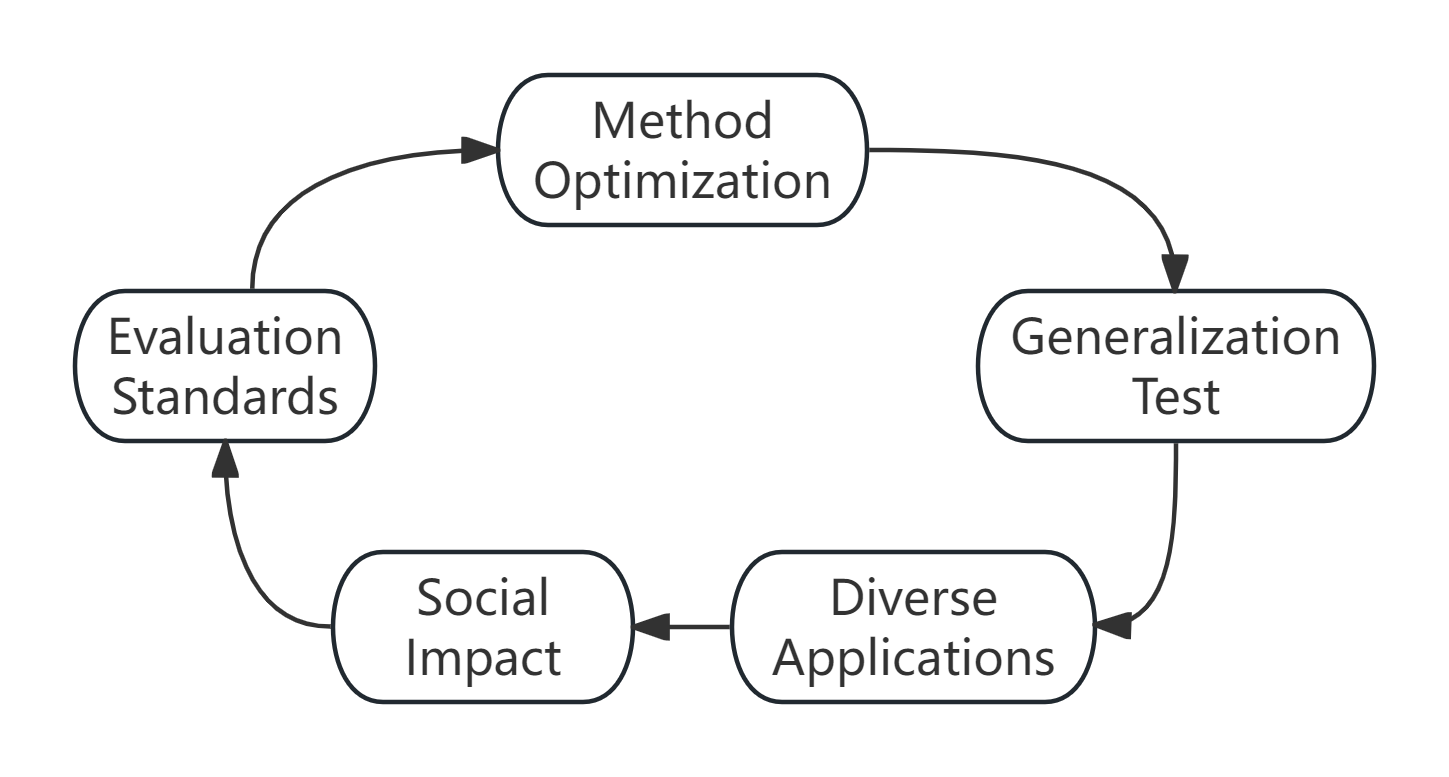}
    \caption{Future agenda of text-to-3D development.}
    \label{fig:discussion}
\end{figure}

In this section, we discuss what might need to be done for the future agenda of text-to-3D. There are five components that are worth attention in a circular manner, as shown in Fig.~\ref{fig:discussion}.

\subsection{\textbf{Evaluation Standards}}

Without objective and human-aligned metrics, it is challenging to accurately measure the performance enhancements of the models and to clearly understand which methods are effective and which areas require further improvement.
Existing evaluation methods for text-to-3D generation often lack reliable automated metrics, focusing mainly on specific criteria like the alignment between generated 3D assets and input text. However, these metrics are inflexible and often misaligned with human preferences. 
To improve this, two notable approaches have emerged~\cite{he2023t, wu2024gpt}: one introduces a comprehensive benchmark with increasing text complexity levels~\cite{he2023t}, 
and the other proposes an automated evaluation method using prompt generation and pairwise comparisons to assign Elo ratings~\cite{wu2024gpt}. 
The first approach evaluates model performance through multi-view quality detection and text alignment, while the second ensures strong alignment with human preferences. With the introduction and discussion of these frameworks, the evaluation methods are gradually evolving towards more objective, adaptable, and human-aligned approaches, laying the groundwork for large-scale, precise text-to-3D evaluations in the future.

\subsection{\textbf{Method Optimization}}

The current landscape of text-to-3D generation faces significant limitations that constrain the production of high-quality and diverse 3D assets. A primary issue is the trade-off between fidelity and speed~\cite{xu2022dream3d,lin2022magic3d}; faster inference often compromises accuracy and detail. Inference speed is a major bottleneck in the text-to-3D process. Additionally, the Janus problem~\cite{hong2023debiasing} affects 2D models' 3D perception, leading to distortion and ghost images, particularly in front-view geometry, which undermines realism.
Controllability~\cite{li2023mvcontrol,chen2023control3d,zhang2023adding} is another concern, as current methods often experience guiding collapse, making it challenging to accurately map object semantics to 3D structures. Furthermore, the lack of precise control based on user prompts limits the versatility and practicality of these models~\cite{zhuang2023dreameditor,fang2023gaussianeditor}. Lastly, limited sample diversity reduces the range of 3D assets generated~\cite{yan2024flow}, curtailing the creative potential of text-to-3D systems.
By addressing these challenges—balancing fidelity and speed, enhancing controllability, solving perception issues, and increasing diversity—the text-to-3D generation field can be more robust, efficient, and diverse.

\subsection{\textbf{Generalization Test}}

With an established evaluation framework and substantial performance improvements, the next step involves investigating its generalization capability. Specifically, the concern is whether the methods developed in controlled laboratory environments can be generalized to the cases in the wild, which is essential for advancing its practical deployment.
Text-to-3D technology, while promising, faces significant challenges that hinder its direct application in traditional 3D scenarios. One major issue is the difficulty in converting text descriptions into 3D representations that can be directly applied. To overcome this barrier, future research may focus on improving representation techniques. Ongoing research efforts, such as \textit{Fantasia3D}~\cite{chen2023fantasia3d}, which separates geometry and appearance modeling to achieve photorealistic rendering, and \textit{CraftsMan}~\cite{li2024craftsman}, which utilizes a 3D diffusion model to generate high-fidelity geometries, are paving the way for more effective applications. Additionally, \textit{VPP}~\cite{qi2024vpp} introduces a progressive generation method that efficiently generates multi-category 3D shapes, addressing current limitations.

\subsection{\textbf{Diverse Applications}}

Upon validating its applicability in the wild, further discussion can focus on the future prospects and innovative directions of text-to-3D technology. This section will examine the potential value of the technology across various domains, highlighting future research directions such as virtual character creation, texture generation, and scene construction. 
Multiple specific text-to-3D application scenarios, such as text-to-avatar, text-to-texture, and text-to-scene, merit further research~\cite{gao2023textdeformer,lin2023componerf,zhuang2023dreameditor,cao2023dreamavatar,song2023roomdreamer,chen2022tango}. Text-to-3D technology has gained attention in digital character creation, particularly in the film and game industries~\cite{lee2021all,yang2022fusing}. Creating digital characters for virtual worlds requires customization based on identities and applying artistic styles or animations through simple motion controls~\cite{wang2023survey,kye2021educational}.
Research on text-to-texture has also gained prominence, as generating textures automatically is challenging due to the need to consider surface shapes~\cite{chen2023text2tex,richardson2023texture,cao2023texfusion}. Textures must align with text prompts while maintaining high quality and consistency on target meshes. Additionally, generating 3D scenes presents challenges like large-scale reconstruction, multi-view images, and ensuring realism and consistency~\cite{song2023roomdreamer}.
Text-driven 3D editing has become a key focus, enabling convenient operations like texture editing, shape deformation, scene decomposition, and stylization through text input. Future research in text-to-3D is expected to explore diverse application directions, addressing current challenges and developing innovative technologies. These advancements could enhance digital content creation tools and expand applications across various industries.

\subsection{\textbf{Social Impact}}

The potential safety and ethical issues arising from the large-scale deployment of this technology need to be considered. It is critical to address any potential negative impacts as the technology advances and becomes more widely adopted, ensuring its use aligns with societal and ethical standards. The application of text-to-3D technology raises security and ethical concerns~\cite{zohny2023ethics,wach2023dark}. For instance, the ability to generate 3D content can be exploited to create deceptive or fabricated information, particularly in contexts like virtual reality, advertising, and media, where misleading representations of individuals or scenarios might deceive the public and undermine societal trust~\cite{lv2023generative}. There is also a risk that this technology could be misappropriated to produce violent, explicit, or otherwise objectionable content, posing significant harm, especially in sensitive environments or to younger audiences. Furthermore, if the generative models are trained on biased datasets, the 3D content produced could unintentionally reinforce societal or cultural biases, resulting in discriminatory or stereotypical representations that exacerbate social inequalities. Notably, such social impact should also reflected in the evaluation metrics in the next stage such that it constitutes a circular and iterative process to develop responsible text-to-3D.

\section{Conclusion}~\label{sec:conclusion}

This review article provides a comprehensive examination of the current state of text-to-3D technologies, covering foundational technologies, the latest advancements, challenges faced, and diverse applications. Specifically, it explores 3D data representation, distinguishing between Structured data (such as voxel grids and multi-view images) and non-structured data (such as meshes, point clouds, and neural fields). The article introduces multiple foundational technologies, including Neural Radiance Field, Diffusion models, Contrastive Language-Image Pre-training, Score Distillation Sampling. It also covers the seminal text-to-3D methods and discusses various directions to address the remaining challenges, such as fidelity, efficiency, consistency, controllability, diversity. Moreover, the work showcases various text-to-3D applications, including text-guided 3D avatar generation, 3D texture generation, 3D scene generation, and 3D editing. Additionally, we discuss its future agenda, including evaluation standards, method optimization, generalization test, diverse applications, and its underlying social impact. Overall, our work contributes to helping readers interested in text-to-3D quickly catch up with its rapid development.

\bibliographystyle{IEEEtran}
\bibliography{local}

\vfill

\end{document}